\documentclass[final]{cvpr}
\usepackage{nopageno}

\usepackage{times}
\usepackage{epsfig}
\usepackage{graphicx}
\usepackage{amsmath}
\usepackage{amssymb}

\usepackage{pifont}
\usepackage{multirow}
\usepackage{diagbox}
\usepackage[skip=4pt]{caption}
\usepackage{tabularx}
\usepackage{balance}
\usepackage{booktabs}

\usepackage[pagebackref=true,breaklinks=true,colorlinks,bookmarks=false]{hyperref}

\usepackage[hang,flushmargin]{footmisc} 

\def\confYear{CVPR 2021}

\renewcommand{\etal}{~\emph{et al.}}
\newcommand{\skipconv}{Skip-Conv~}
\newcommand{\skipconvdot}{Skip-Conv}

\begin{document}

\title{Skip-Convolutions for Efficient Video Processing\vspace{-0.5em}}
\author{
Amirhossein Habibian \qquad Davide Abati \qquad Taco S. Cohen \qquad Babak Ehteshami Bejnordi\vspace{0.5em}\\
Qualcomm AI Research\thanks{Qualcomm AI Research is an initiative of Qualcomm Technologies, Inc.}\\
\small{\texttt{\{habibian,dabati,tacos,behtesha\}@qti.qualcomm.com}}
}

\maketitle

\begin{abstract}
We propose Skip-Convolutions to leverage the large amount of redundancies in video streams and save computations.
Each video is represented as a series of changes across frames and network activations, denoted as residuals.
We reformulate standard convolution to be efficiently computed on residual frames:
each layer is coupled with a binary gate deciding whether a residual is important to the model prediction,~\eg foreground regions, or it can be safely skipped,~\eg background regions. 
These gates can either be implemented as an efficient network trained jointly with convolution kernels, or can simply skip the residuals based on their magnitude.
Gating functions can also incorporate block-wise sparsity structures, as required for efficient implementation on hardware platforms.
By replacing all convolutions with Skip-Convolutions in two state-of-the-art architectures, namely EfficientDet and HRNet, we reduce their computational cost consistently by a factor of $3\sim4\times$ for two different tasks, without any accuracy drop.
Extensive comparisons with existing model compression, as well as image and video efficiency methods demonstrate that Skip-Convolutions set a new state-of-the-art by effectively exploiting the temporal redundancies in videos.
\end{abstract}
\setlength{\parindent}{0pt}
\section{Introduction}
\label{sec:introduction}
Is a video a sequence of still images or a continuous series of changes?
We see the world by sensing changes, and process information whenever the accumulated differences in our neurons exceed some threshold.
This trait has inspired many efforts to develop neuromorphic sensors and processing algorithms, such as event-based cameras~\cite{posch2010high} and spiking neural networks~\cite{gerstner2002spiking}.
Despite their efficiency for video processing, spiking nets have not been as successful as conventional models, mostly due to the lack of efficient training algorithms. There have been several works on mapping spiking nets to conventional networks, but these works have been mostly limited to shallow architectures and simple problems, such as digit classification~\cite{zambrano2016fast,o2016sigma,o2018temporally}. Representing videos by changes through residual frames is common in video compression codecs, such as HEVC~\cite{sullivan2012overview}, because residual frames normally have less information entropy and therefore require fewer bits to be compressed.
\begin{figure}[t]
\centering
\includegraphics[width=0.99\columnwidth]{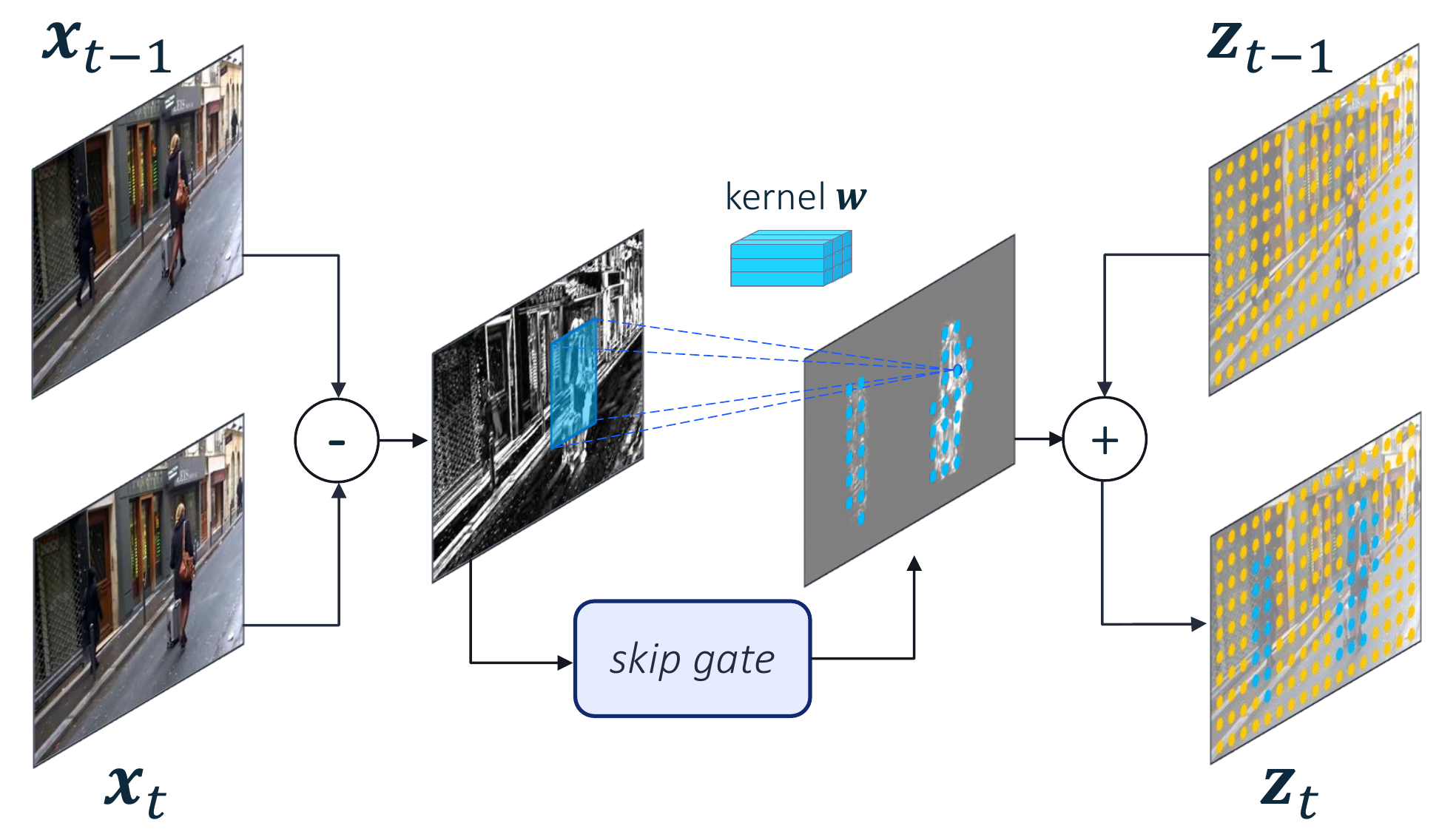}
\vspace{-2mm}
\caption{
Skip-Convolution illustration for the input layer.
Convolutions are computed only on a few locations in the residual features determined by a gate function (blue dots).
In other locations, output features are copied from the previous time step (yellow dots).
Frames taken from ~\cite{cc1}.
}
\label{fig:cover}
\vspace{-3mm}
\end{figure}

For stream processing applications, that require spatially dense predictions for each input frame, deep convolutional networks still process a sequence of still images as input.
Each frame is processed entirely by sliding convolutional filters all over the frame, layer by layer.
As a result, the overall computational cost grows linearly with the number of input frames, even though there might be not much new information in the subsequent frames.
This inherent inefficiency prohibits using accurate but expensive networks for real-time tasks, such as object detection and pose estimation, on video streams.

This paper proposes \textit{Skip-Convolutions}, in short, \skipconvdot s, to speed up any convolutional network for inference on video streams. Instead of considering a video as a sequence of still images, we represent it as a series of changes across frames and network activations, denoted as residual frames.
We reformulate standard convolution to be efficiently computed over such residual frames by limiting the computation only to the regions with significant changes while skipping the others.
Each convolutional layer is coupled with a gating function learned to distinguish between the residuals that are important for the model accuracy and background regions that can be safely ignored (Fig.~\ref{fig:cover}).

By applying the convolution kernel on sparse locations, \skipconvdot s allow to adjust efficiency depending on the input, in line with recent studies on conditional computation in images~\cite{li2017not, dong2017more, ren2018sbnet, verelst2020dynamic, wang2020learning}. However, we hereby argue that distinguishing the important and non-important regions is more challenging in still images. Indeed, residual frames provide a strong prior on the relevant regions, easing the design of effective gating functions. As a result, \skipconvdot s achieve a much higher cost reduction in videos ($300\sim400 \%$), compared to what has been previously reported for images ($15\sim60\%$ in~\cite{verelst2020dynamic}, $27\sim41\%$ in~\cite{wang2020learning}).

To summarize, the main contributions of this work are: \emph{i)} a simple reformulation of convolution, which computes features on highly sparse residuals instead of dense video frames. \emph{ii, iii)} Two gating functions, Norm gate and Gumbel gate, to effectively decide whether to process or skip each location. Norm gates do not have any trainable parameter, thus can be easily plugged into any trained network obviating the need for further fine-tuning. On the contrary, Gumbel gates are trainable: they are learned jointly with the backbone model with the Gumbel reparametrization~\cite{jang2016categorical,maddison2016concrete}, and allow to achieve even more efficiency.
We extend these gates to generate structured sparsity as required for efficient hardware implementations. \emph{iv)} A general formulation of~\skipconvdot, which extends the idea to a broader range of transformations and operations.
\emph{v)} extensive experiments on two different tasks and state-of-the-art network architectures, showing a consistent reduction in cost by a factor of $3\sim4\times$, without any accuracy drop.
\section{Related work}
\label{sec:related_work}
\paragraph{Efficient video models}
Exploiting temporal redundancy is the key to develop efficient video models. A common strategy is feature propagation~\cite{shelhamer2016clockwork, zhu2017deep,li2018low, zhu2018towards}, which computes the expensive backbone features only on key-frames. 
Subsequent frames then adapt the backbone features from key-frames directly~\cite{shelhamer2016clockwork} or after spatial alignments via optical flow~\cite{zhu2017deep,jain2019accel}, dynamic filters~\cite{li2018low,nie2019dynamic}, or self-attention~\cite{hu2020temporally}. Similarly, ~\skipconv also propagates features from the previous frame, however: \emph{i)} feature propagation models depend on the alignment step, which is potentially expensive,~\eg for accurate optical flow extraction. \emph{ii)} These methods propagate the feature only at a single layer, whereas~\skipconv propagates features at every layer. \emph{iii)} \skipconv selectively decides whether to propagate or compute at the pixel level, rather than for the whole frame. \emph{iv)} differently  from feature propagation methods that imply architectural adjustments, ~\skipconv does not involve any modifications to the original network.

Another strategy is to interleave deep and shallow backbones between consecutive frames~\cite{jain2019accel,liu2019looking,nie2019dynamic}. The deep features, extracted only on key-frames, are fused with shallow features extracted on other frames using concatenation~\cite{jain2019accel}, recurrent networks~\cite{liu2019looking}, or more sophisticated dynamic kernel distillation~\cite{nie2019dynamic}. This strategy usually leads to an accuracy gap between key-frames and other frames.

Several works aim for efficient video classification by developing faster alternatives for 3D convolutions, such as temporal shift modules~\cite{lin2019tsm} and 2+1D convolutions~\cite{tran2018closer}, neural architecture search~\cite{piergiovanni2019tiny,feichtenhofer2020x3d}, or adaptive frame sampling~\cite{campos2018skip,wu2019liteeval,meng2020ar}. These methods are mostly suitable for global prediction tasks where a single prediction is made for the whole clip. Differently, we target stream processing tasks, such as pose estimation and object detection, where a spatially dense prediction is required for every frame.

\paragraph{Efficient image models}
The reduction of parameter redundancies,~\eg in channels and layers, is a fundamental aspect for obtaining efficient image models.
Model compression methods~\cite{kuzmin2019taxonomy}, such as low-rank tensor decomposition~\cite{jaderberg2014speeding,wsvd}, channel pruning~\cite{he2017channel, louizos2018learning}, neural architecture search~\cite{tan2019efficientnet,tan2020efficientdet}, and knowledge distillation~\cite{hinton2015distilling, romero2014fitnets}, effectively reduce the memory and computational cost of any network. Instead of exploiting weight redundancies, as addressed by model compression,~\skipconv leverages temporal redundancies in activations. As verified by our experiments, these are complementary and can be combined to further reduce the computational cost.

Conditional computation has recently shown great promise to develop efficient models for images~\cite{bengio2015conditional}.
It enables the model to dynamically adapt the computational graph per input to skip processing unnecessary branches~\cite{huang2018multi}, layers~\cite{veit2018convolutional}, channels~\cite{bejnordi2019batch, gao2018dynamic}, or non-important spatial locations such as background~\cite{figurnov2017spatially, li2017not, dong2017more, ren2018sbnet, verelst2020dynamic, wang2020learning}. However, distinguishing the important vs. non-important regions is difficult in images.~\skipconv leverages residual frames as a strong prior to identify important regions in feature maps based on their changes, outperforming their image counterparts by a large margin as validated by our experiments.
\section{Skip Convolutions}
\label{sec:method}
\def\z{\mathbf{z}}
\def\ztilde{\tilde{\z}}
\def\zt{\z_t}
\def\ztt{\z_{t-1}}
\def\x{\mathbf{x}}
\def\xt{\x_t}
\def\xtt{\x_{t-1}}
\def\r{\mathbf{r}}
\def\rt{\r_t}
\def\rtt{\r_{t-1}}
\def\g{g}
\def\gl{g_l}
\def\f{f}
\def\phil{\phi_l}
\def\w{\mathbf{w}}
\def\R{\mathbb{R}}
\newcommand{\norm}[1]{\left\lVert#1\right\rVert}
\newcommand\round[1]{\text{round}\left(#1\right)}
Instead of treating a video as a sequence of still images, we represent it as a series of residual frames defined both for the input frames and for intermediate feature maps. In section~\ref{sec:method_residual_domain}, we reformulate the standard convolution to be efficiently computed on residuals.
Section~\ref{sec:method_gating} proposes several gating functions to decide whether to process or skip each location in residual frames. Gating functions are crucial to reduce the computation without losing much accuracy.
Finally, section~\ref{sec:method_generalized} discusses how \skipconv can be generalized to a broader set of transformations beyond residuals as a direction for future developments.
\subsection{Convolution on Residual Frames}
\label{sec:method_residual_domain}
\begin{figure}[tb]
\bgroup
\setlength{\tabcolsep}{1pt}
\centering
\resizebox{0.95\columnwidth}{!}{
\begin{tabular}{ccc}
\includegraphics[width=0.33\columnwidth]{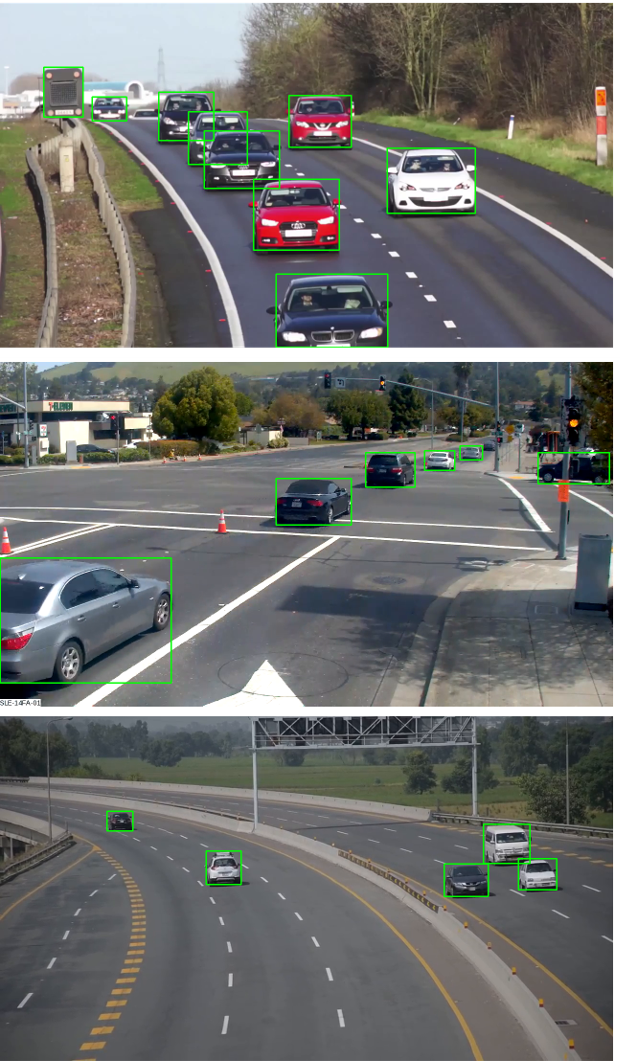}&
\includegraphics[width=0.33\columnwidth]{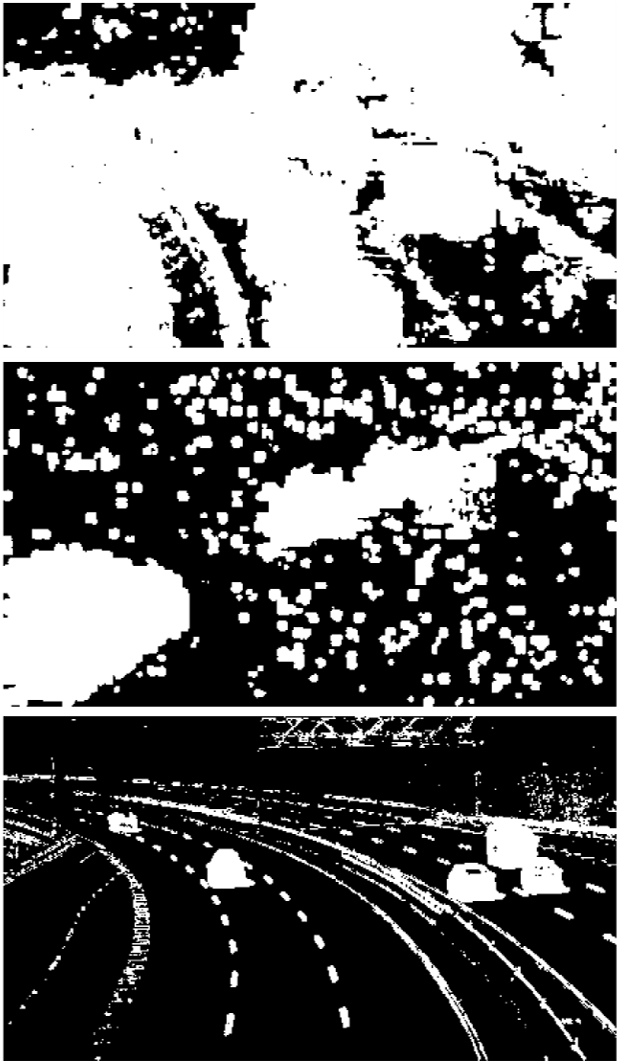}&
\includegraphics[width=0.33\columnwidth]{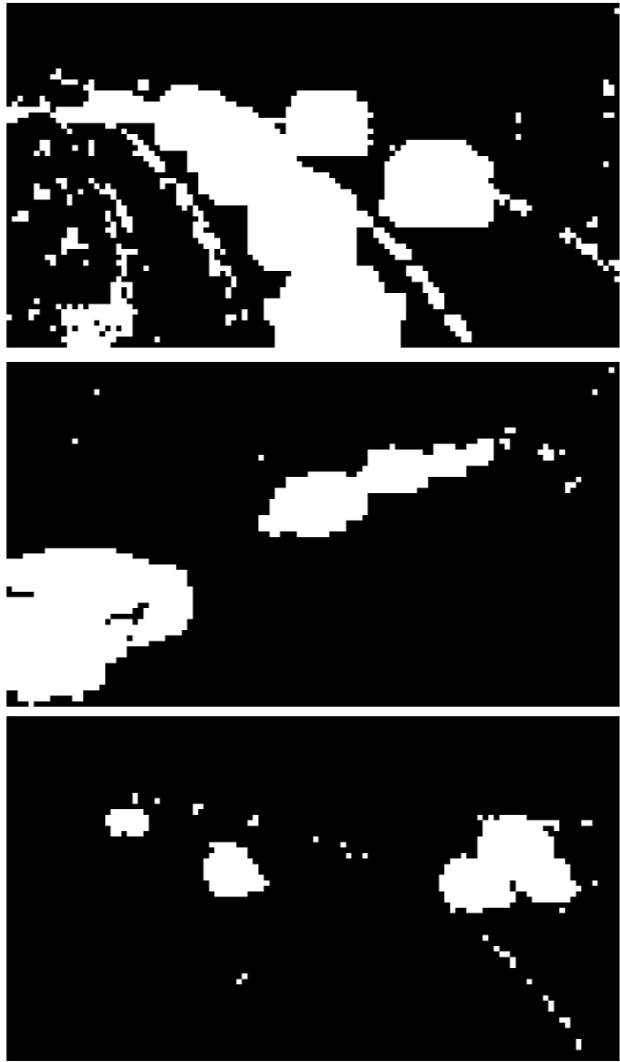}\\
Input frame & Layer 3 & Layer 30 
\end{tabular}}
\vspace{-2mm}
\caption{
Gating masks for video object detection.
Gates become more selective at deeper layers, concentrating on task specific regions.
Frames from~\cite{cc2,cc4,cc3}.
}
\label{fig:object_detection_layers}
\egroup
\vspace{-4mm}
\end{figure}

Given a convolutional layer with a kernel $\w \in \R^{c_o \times c_i \times k_h \times k_w}$ and an input $\xt \in \R^{c_i \times h \times w}$, the output feature map $\zt \in \R^{c_o \times h \times w}$ is computed for each frame as\footnote{To avoid notational clutter, we describe the case in which $\xt$ and $\zt$ have the same resolution.}:
\begin{equation}
\label{eq:conv}
\zt = \w \ast \xt.
\end{equation}
In Eq.~\ref{eq:conv} (and in the remainder of this section) $\zt$ refers to the result before the application of a non-linear activation function.
Using the distributive property of convolution as a linear function, the output can be obtained by:
\begin{align}
\label{eq:fact_t}
\begin{split}
\zt 
&= \w \ast \xtt + \w \ast \xt - \w \ast \xtt \\
&= \ztt + \w \ast (\xt - \xtt)\\
&= \ztt + \w \ast \rt,
\end{split}
\end{align}
where $\rt$ represents the residual frame as the difference between the current and previous feature maps $\xt - \xtt$. Since $\ztt$ has been already computed for the previous frame, computing $\zt$ reduces to summing the term $\w \ast \rt$.
Due to the high correlation of consecutive frames in a video, the residual frame $\rt$ is often sparse and contains non-zero values only for the regions that changed across time,~\ie moving objects as visualized in Figure~\ref{fig:object_detection_layers}. This sparsity can effectively be leveraged for efficiency: for every kernel support filled with zero values in $\rt$, the corresponding output will be trivially zero, and the convolution can be skipped by copying values from $\ztt$ to $\zt$.

We use residuals to represent features at every convolutional layer. For the first frame, the residual $\r_1$ will be the same as the frame content $\x_1$, so the feature map is computed over the whole frame. 
Instead, consecutive frames update their features only at locations with non-zero residuals while reusing past representations elsewhere.

Although residuals are inherently sparse, they may still contain lots of locations with small non-zero values that prevent skipping them. To save even further, we introduce a gating function for each convolutional layer, $\g: \mathbb{R}^{c_i \times h \times w} \rightarrow \{0, 1\}^{h \times w}$, to predict a binary mask indicating which locations should be processed, and taking only $\r_t$ as input. Using $\r_t$ as input provides a strong prior to the gating function, making it effective even with a fairly simple form. Putting it all together, our proposed~\skipconv is defined as:
\begin{equation}
\label{eq:skipconv}
\ztilde_t = \ztilde_{t-1} + \g(\rt) \odot (\w \ast \rt),
\end{equation}
where $\odot$ indicates a broadcasted Hadamard (i.e. elementwise) product and the $\sim$ symbol highlights that $\ztilde_t$ is an approximation of $\zt$, as it skips negligible but non-zero residuals. The gating function is further described next.
\subsection{Skipping Non-zero Residuals}
\label{sec:method_gating}
We propose two gating functions: \emph{i)} Norm gate, that decides to skip a residual if its magnitude (norm) is small enough. This gate does not have any learnable parameter and does not involve any training. As a result, it can be easily plugged into any trained image network without any labeled video or training resources required. \emph{ii)} Gumbel gate, that has parameters trained jointly with the convolutional kernels. The learned parameters can make the Gumbel gate more effective at the cost of fine-tuning the model.
\subsubsection{Norm Gate}
A naive form of gating is based on applying a scalar threshold $\epsilon$ to the norm of each output pixel:
\begin{equation}
\label{eq:pt_ideal}
\g(\rt, \w, \epsilon) = \round{\sigma(\norm{\w\ast\rt}_p - \epsilon)},
\end{equation}
where $\sigma(\cdot)$ indicates a sigmoid function, $p$ represents the order of the norm, and the norm is computed over all channels for each position.
However, such a gating function requires the computation of the convolution at each pixel of the residual, which would reintroduce inefficiency.
We therefore propose to approximate Eq.~\ref{eq:pt_ideal} by considering the norm of each kernel support in the residual as:
\begin{equation}
\label{eq:pt_input}
\g(\rt, \epsilon) = \round{\sigma(\norm{\rt}_p - \epsilon)},
\end{equation}
We refer to this function as \textit{Input-Norm gate}.
The norm $\norm{\rt}_p$ in Eq.~\ref{eq:pt_input} is to be intended for local convolutional supports rather than pixel-wise. As such, it is computed by 
applying an absolute value function to $\rt$ then taking sum within the $d_i \times k_h \times k_w$ neighborhood (\ie $p=1$).

A more accurate approximation can be achieved without computing the full convolution, by involving the norm of the weight matrix $\w$. 
Considering Young's inequality~\cite{young1912multiplication} we get an upper bound on the norm of the convolution of two vectors $\textbf{f}$ and $\textbf{g}$:
\begin{align}
\begin{split}
\label{eq:young}
\norm{\textbf{f} \ast \textbf{g}}_r \leq \norm{\textbf{f}}_s \cdot  \norm{\textbf{g}}_q,\\
\text{where}\quad{\frac {1}{s}}+{\frac {1}{q}}={\frac {1}{r}}+1.
\end{split}
\end{align}
By following Eq.~\ref{eq:young}, we define a more precise approximation, based on the norms of the input residual $\rt$ and the weight matrix $\w$, in what we refer to as \textit{Output-Norm gate}:
\begin{equation}
\label{eq:pt_output}
\g(\rt, \w, \epsilon) = \round{\sigma(\norm{\w}_p \cdot\norm{\rt}_p - \epsilon)},
\end{equation}
where the norm $\norm{\w}_p$ is computed over all four dimensions.
We set the order $p$ for both input-norm and output-norm gates to 1 (i.e., $l_1$ norm), and we share the margin $\epsilon$ between all layers.
More flexible strategies such as layer-specific $\epsilon$ can potentially yield better results at the cost of more hyperparameter tweaking.
\subsubsection{Gumbel Gate}
Residual norms indicate regions that change significantly across frames.
However, not all changes are equally important for the final prediction (\eg changes in background).
This observation suggests that a higher efficiency can be gained by introducing some trainable parameters within gates, which are learned to skip even large residuals when they do not affect the model performance.

For each convolutional layer $l$ we define a light-weight gating function $\f(\rt; \phil)$, parameterized by $\phil$, as a convolution with a single output channel. 
Such an addition imposes a negligible overhead to the convolutional layer, which normally has dozen to hundreds of output channels. To generate masks of the same resolution, the gate function uses the same kernel size, stride, padding, and dilation as its corresponding layer.
The gating function $f$ outputs unnormalized scores that we turn into pixel-wise Bernoulli distributions by applying a sigmoid function.
During training, we sample binary decisions from the Bernoulli distribution, whereas we round the sigmoids at inference:
\begin{equation}
\g(\rt,\phil) \begin{cases}
\sim \text{Bern}(\sigma(\f(\rt; \phil))) &\text{at training},\\
= \round{\sigma(\f(\rt; \phil))} &\text{at inference}
\end{cases}
\end{equation}

We employ the Gumbel reparametrization~\cite{jang2016categorical,maddison2016concrete} and a straight-through gradient estimator~\cite{bengio2013estimating} in order to backpropagate through the sampling procedure. The gating parameters are learned jointly with all model parameters by minimizing $\mathcal{L}_{task} + \beta \mathcal{L}_{gate}$. The hyper-parameter $\beta$ balances the model accuracy, measured by $\mathcal{L}_{task}$, vs the model efficiency as measured by $\mathcal{L}_{gate}$. We define the gating loss as the average multiply-accumulate (MAC) count needed to process $T$ consecutive frames as:
\begin{equation}
\mathcal{L}_{gate}(\phi_1,\dots,\phi_L) = \frac{1}{T-1}\sum_{t=2}^T\sum_{l=1}^L m_l \cdot \mathbb{E}[\g(\rt,\phil)],
\end{equation}
where $L$ is the number of layers in the network, $\mathbb{E}[\cdot]$ indicates an average over spatial locations and the coefficient $m_l$ denotes the MAC count for the $l^{th}$ convolutional layer~\footnote{To keep $\mathcal{L}_{gate}$ at a manageable scale, we normalize $m_l$ by dividing it by $\sum_{i=1}^L m_i$.}. 
Similar to recurrent networks, we train the model over a fixed-length sequence of frames and do inference iteratively on an arbitrary number of frames.
\paragraph{Structured Sparsity} 
\begin{figure}[tb]
\bgroup
\setlength{\tabcolsep}{2pt}
\centering
\resizebox{0.95\columnwidth}{!}{
\begin{tabular}{cccc}
\includegraphics[width=0.25\columnwidth]{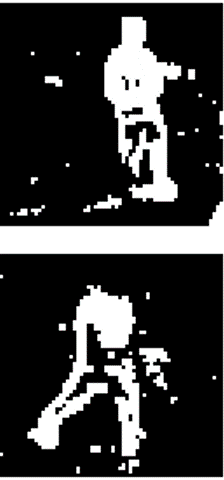}&
\includegraphics[width=0.25\columnwidth]{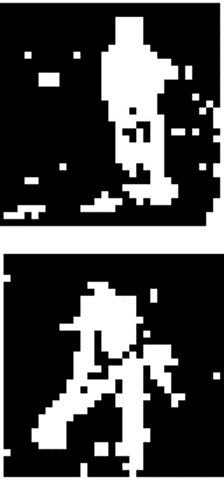}&
\includegraphics[width=0.25\columnwidth]{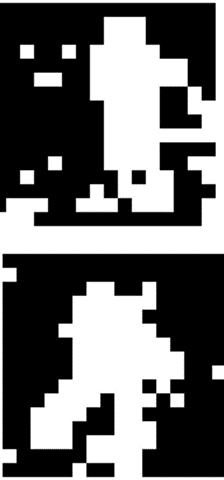}&
\includegraphics[width=0.25\columnwidth]{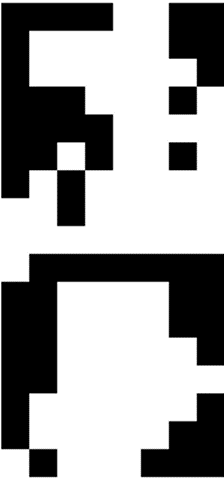}\\
1x1 & 2x2 & 4x4 & 8x8
\end{tabular}}
\vspace{-2mm}
\caption{Exemplar masks generated by \skipconv for pose estimation, when trained with different block structures.}
\label{fig:plot_structure_blocks}
\egroup
\vspace{-3mm}
\end{figure}
Similar to sparse convolutions, an efficient implementation of~\skipconv requires block-wise structured sparsity in the feature maps~\cite{ren2018sbnet, verelst2020dynamic}, for two main reasons. 
First, block structures can be leveraged to reduce the memory overhead involved in gathering and scattering of input and output tensors~\cite{ren2018sbnet}. Additionally, many hardware platforms perform the convolutions distributed over small patches (\eg $8\times8$), so do not leverage any fine-grained spatial sparsity smaller than these block sizes.

\skipconv can be extended to generate structured sparsity by simply adding a downsampling and an upsampling function on the predicted gates. More specifically, we add a max-pooling layer with the kernel size and stride of $b$ followed by a nearest neighbor upsampling with the same scale factor of $b$.
This enforces the predicted gates to have $b \times b$ structure, as illustrated in Figure~\ref{fig:plot_structure_blocks}.
We will illustrate in Section~\ref{sec:ablation} how adding structure, despite significantly reducing the resolution of the gates, does not harm performances when compared with unstructured gating.
Thus, structured sparsity enables more efficient implementation with minimal effect on performance.
\subsection{Generalization \& Future Work} 
\label{sec:method_generalized}
\skipconv computes the output features in three steps: \emph{i)} encoding the input tensor as residuals using a global subtraction transform. \emph{ii)} efficient computation in the residual domain by leveraging the sparsity. \emph{iii)} decoding the output back into the feature space using a global addition transform (inverse of subtraction). 
Here we generalize this process to a broader set of transformations beyond global subtraction/additions for the interested reader, but leave these ideas to be explored in future work.

Whereas in Eq.~\ref{eq:fact_t} we defined the residual as $\rt = \xt - \xtt$, we may more generally define 
$\rt = f_{\xtt}(\xt)$ as an $\xtt$-dependent (approximately) invertible function $f$ of $\xt$, that produces a sparse generalized residual $\rt$.
As before, we may then write:
\begin{align}
\label{eq:fact_t_generalized}
\begin{split}
\zt 
= \w \ast \xt = 
\w \ast f_{\xtt}^{-1}(\rt).
\end{split}
\end{align}

Now, if convolution is equivariant to $f_{\xtt}^{-1}$, i.e. if the equation $\w \ast \f_{\xtt}^{-1}(\rt) = \tilde{f}_{\xtt}^{-1}(\w \ast \rt)$ holds for some function $\tilde{f}_{\xtt}^{-1}$ acting on the output space of the convolution, then we can compute $\zt$ via a convolution with a sparse $\rt$ followed by a transformation by $\tilde{f}$ (which should be chosen to be efficiently computable):
\begin{equation}
    \zt = \w \ast f_{\xtt}^{-1}(\rt) = \tilde{f}_{\xtt}^{-1}(\w \ast \rt).
\end{equation}
The original \skipconv is recovered by setting $f_{\xtt}(\xt) = \xt - \xtt$, so that $f^{-1}_{\xtt}(\rt) = \rt + \xtt$ and the output transformation is $\tilde{f}_{\xtt}^{-1}(\w \ast \rt) = \ztt +\w \ast \rt$.

The question of when a convolution is equivariant to a given group of transformations has received a lot of attention in the literature \cite{Cohen2016group,Kondor2018generalization,Cohen2019general}.
The general answer is that $\w \ast$ can be made equivariant by linearly constraining the filters, resulting in so-called steerable filters \cite{Freeman1991design}.
In this case, however, the group of transformations generated by $f_{\xtt}$ for all $\xtt$ may not be known in advance, so analytically solving the linear constraints on the filters is not feasible.
Nevertheless, equivariance can be encouraged via a simple loss term that pulls $\w \ast f_{\xtt}^{-1}(\rt)$ and $\tilde{f}^{-1}_{\xtt}(\w \ast \rt)$ closer.

One promising choice for $f_{\xtt}$ is to compute a residual between $\xt$ and a \emph{warped} version of $\xtt$.
This operation is guaranteed to be invertible (just add back the warped $\xtt$) and is equivariant whenever the warping operation is equivariant. 
Formally, let $T$ denote a warping operation,~\eg bilinear interpolation of a frame at a set of points indicated by a flow field.
The flow field could be computed from the network input frames, for instance.
We may define $f_{\xtt}(\xt) = \xt - T(\xtt)$, so that $f_{\xtt}^{-1}(\rt) = \rt + T(\xtt)$ and, if the warp is equivariant, $\tilde{f}_{\xtt}^{-1}(\w \ast \rt) = \tilde{T}(\ztt) + \w \ast \rt$ (where $\tilde{T}$ applies the warp to the convolution output space).
Other choices for the function $f$ and $\tilde{f}$ also apply, including learning them from data.
\section{Experiments}
\label{sec:experiments}
We evaluate~\skipconv on two stream processing tasks, namely object detection and single-person pose estimation, in Section~\ref{sec:video_object_detection} and~\ref{sec:human_pose_estimation} respectively. Several ablation studies are reported in Section~\ref{sec:ablation}.
\subsection{Object Detection}
\label{sec:video_object_detection}
\paragraph{Experimental setup}
We conduct object detection experiments on UA-DETRAC dataset~\cite{detrac}. It consists of over 140,000 frames capturing 100 real-world traffic videos with bounding box annotations provided for vehicles at every frame. The dataset comes with a standard partitioning of $60$ and $40$ videos as train and test data, respectively.
The performance is evaluated in terms of average precision (AP), averaged over multiple IoU thresholds varying from 0.5 to 0.95 with a step size of 0.05, similar to~\cite{tan2020efficientdet}.
\paragraph{Implementation details}
We use EfficientDet~\cite{tan2020efficientdet}, the state of the art architecture for object detection, and apply~\skipconv on top of it. We conduct our experiments on D0 to D3 as the most efficient configurations~\cite{tan2020efficientdet}, though more expensive configurations,~\ie D4 to D7, can similarly benefit from~\skipconvdot.
Each model is initialized with pre-trained weights from MS COCO dataset~\cite{lin2014microsoft} and trained using SGD optimizer with momentum $0.9$, weight decay $4e-5$ and an initial learning rate of $0.01$ for $4$ epochs. We decay the learning rate of a factor of $10$ at epoch $3$. All models are trained with mini-batches of size $4$ using four GPUs, where synchronized batch-norm is used to handle small effective batch sizes.
We use \skipconv with learned gates, which is trained for each EfficientDet configuration using the sparsity loss coefficient set to $\beta=0.01$. 
During training we apply random flipping as data augmentation. The clip length is set to $4$ frames both for training and inference. 
\begin{figure}[tb]
\centering
\includegraphics[width=0.9\columnwidth]{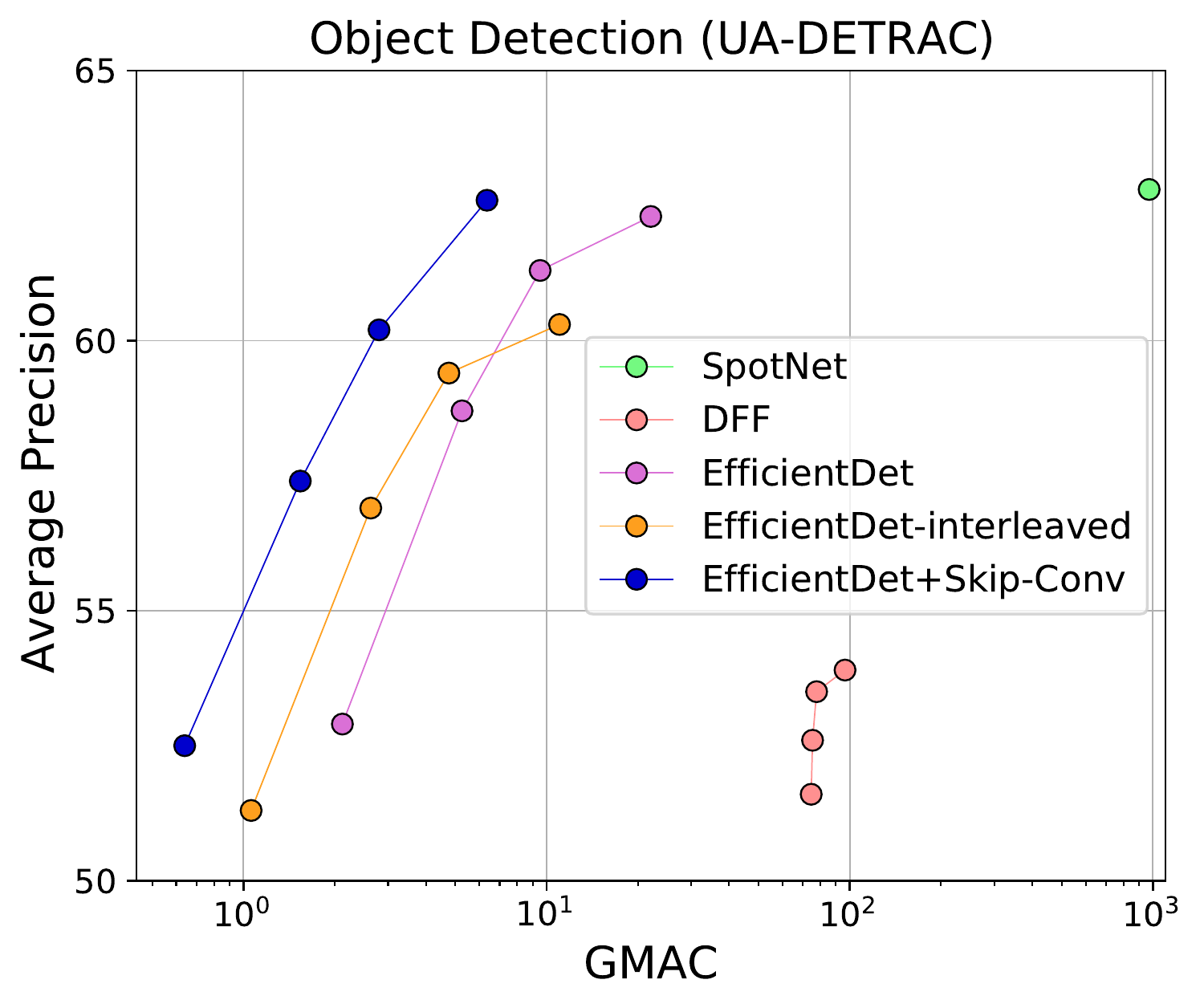}
\vspace{-2mm}
\caption{Comparison with the video object detection state-of-the-art. \skipconv reduces EfficientDet cost by $300\%$, consistently across the configurations D0, D1, D2, D3.}
\label{fig:plot_detrac}
\vspace{-3mm}
\end{figure}
\paragraph{Comparison to state of the art}
We compare~\skipconv to several image and video object detectors: \emph{i)} EfficientDet~\cite{tan2020efficientdet} as the state of the art in efficient object detection in images. We also include an EfficientDet-interleaved baseline, where model predictions are propagated from keyframes to the next frames without further processing. \emph{ii)} Deep Feature Flow (DFF)~\cite{zhu2017deep} as a seminal work on efficient object detection in video, \emph{iii)} SpotNet~\cite{perreault2020spotnet} as the top performer in UA-DETRAC benchmark, which trains a joint model to detect objects and extract motion masks for improved object detection in video.
Figure~\ref{fig:plot_detrac} demonstrates that~\skipconv significantly reduces the computational cost of EfficientDet with a reasonable accuracy drop. More specifically, for D3 configuration, \skipconv reduces the cost from $22.06$ to $6.36$ GMAC with even a slight increase in AP from $62.3$ to $62.6$. Similarly for other configurations, ~\skipconv consistently reduces the MAC count by $330\%$ to $350\%$.
By comparing ~\skipconv and EfficientDet-interleaved, we observe that although interleaved detection reduces the computational cost, it leads to severe accuracy drop as there are lots of motion and dynamics in this dataset. 

Moreover, we observe that~\skipconv outperforms DFF~\cite{zhu2017deep} both in terms of accuracy and computational cost. 
We hypothesize that DFF performances, solely relying on optical-flow to warp features across frames, are sensitive to the accuracy of the predicted motion vectors.
However, there are lots of small objects (\eg distant vehicles) in this dataset for which optical flow predictions are noisy and inaccurate. 
Finally, our experiments demonstrate that ~\skipconv achieves the state of the art accuracy on UA-DETRAC dataset, reported by SpotNet~\cite{perreault2020spotnet}, with orders of magnitude less computes ($6.36$ vs $972.0$ GMAC).
\subsection{Human Pose Estimation}
\label{sec:human_pose_estimation}
\begin{table}[tb]
\centering
\resizebox{\columnwidth}{!}{
\begin{tabular}{lccccccccc}
\toprule
& \textbf{GMAC} & \textbf{Head} & \textbf{Sho}. & \textbf{Elb}. & \textbf{Wri}. & \textbf{Hip} & \textbf{Knee} & \textbf{Ank}. & \textbf{Avg}\\
\midrule
Park\etal~\cite{park2011n} & - & 79.0 & 60.3 & 28.7 & 16.0 & 74.8 & 59.2 & 49.3 & 52.5\\
Nie\etal~\cite{xiaohan2015joint} & - & 80.3 & 63.5 & 32.5 & 21.6 & 76.3 & 62.7 & 53.1 & 55.7\\
Iqbal\etal~\cite{iqbal2017pose} & - & 90.3 & 76.9 & 59.3 & 55.0 & 85.9 & 76.4 & 73.0 & 73.8\\
Song\etal~\cite{song2017thin} & - & 97.1 & 95.7 & 87.5 & 81.6 & 98.0 & 92.7 & 89.8 & 92.1\\
Luo\etal~\cite{luo2018lstm} & 70.98 & 98.2 & 96.5 & 89.6 & 86.0 & 98.7 & 95.6 & 90.9 & 93.6\\
DKD\etal~\cite{nie2019dynamic} & 8.65 & 98.3 & 96.6 & 90.4 & 87.1 & 99.1 & 96.0 & \textbf{92.9} & 94.0\\
\midrule
HRNet-w32~\cite{hrnet} & 10.19 & 98.5 & 97.3 & 91.8 & 87.6 & 98.4 & 95.4 & 90.7 & 94.5\\
\quad+S-SVD~\cite{jaderberg2014speeding} & 5.04 & 97.9 & 96.9 & 90.6 & 87.3 & 98.7 & 95.3 & 91.1 & 94.3\\
\quad+W-SVD~\cite{wsvd} & 5.08 & 97.9 & 96.3 & 87.2 & 82.8 & 98.1 & 93.2 & 88.8 & 92.4\\
\quad+L0~\cite{louizos2018learning} & 4.57 & 97.1 & 95.5 & 86.5 & 81.7 & 98.5 & 92.9 & 88.6 & 92.1\\
\midrule
\quad\textbf{+\skipconv} & 5.30 & \textbf{98.7} & \textbf{97.7} & \textbf{92.0} & \textbf{88.1} & \textbf{99.3} & \textbf{96.6} & 91.0 & \textbf{95.1}\\
\bottomrule
\end{tabular}}
\caption{Comparison with the state-of-the-art on JHMDB. \skipconv outperforms in PCK the best image and video models, whilst requiring fewer MAC per frame.}
\label{tab:pose_estimation_sota}
\vspace{-5mm}
\end{table}%
\paragraph{Experimental setup}
We conduct our experiments on the JHMDB dataset~\cite{jhuang2013towards}, a collection of 11,200 frames from 316 video clips, labeled with 15 body joints.
Video sequences are organized according to three standard train/test partitions and we report average results over the three splits. 
We evaluate the performance using the standard PCK metric~\cite{yang2012articulated}. Given a bounding box of the person with height $h$ and width $w$, PCK considers a candidate keypoint to be a valid match if its distance with the ground-truth keypoint is lower than $\alpha\cdot\max(h, w)$. We set $\alpha=0.2$. Our experimental setup is consistent with prior works~\cite{song2017thin,luo2018lstm,nie2019dynamic}.

\paragraph{Implementation details}
We use HRNet~\cite{hrnet}, the state of the art architecture for human pose estimation, and apply~\skipconv on top of it. We select HRNet-w32 as it performs on par with HRNet-w48, while being more efficient. All models are trained for $100$ epochs with mini-batches of $16$ images, using the Adam optimizer~\cite{adam} with an initial learning rate of $0.001$. We decay the learning rate with a factor of $10$ at epochs 40 and 80.
We use \skipconv with learned gates, which is trained using the sparsity loss coefficient set to $\beta=1e-5$ unless specified otherwise.

We follow the setup from~\cite{song2017thin,luo2018lstm,nie2019dynamic} for training and inference.
We use standard data augmentations during training: randomly scaling using a factor within $[0.6, 1.4]$, random rotation within $[-40^{\circ}, 40^{\circ}]$ and random flipping. Each frame is cropped based on the ground-truth bounding box and padded to $256\times256$ pixels. The inference is done on a single scale.
The clip length is set to $T=8$ frames both for training and inference. 
\begin{figure}[tb]
\centering
\includegraphics[width=0.9\columnwidth]{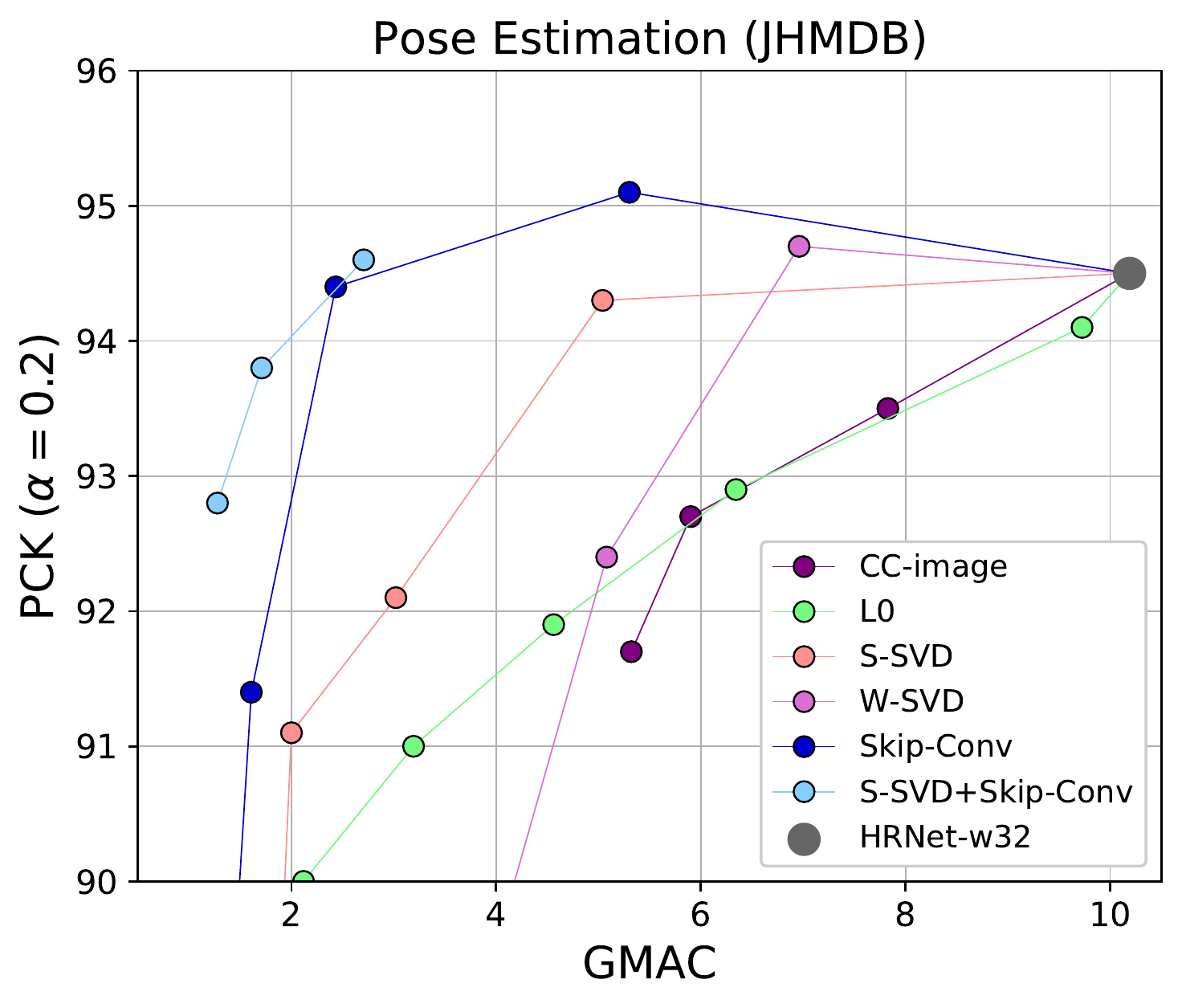}
\vspace{-2mm}
\caption{
Comparison with model compression on JHMDB. \skipconv outperforms existing approaches. Applying it on top of compressed models further improves efficiency.
}
\label{fig:plot_jhmdb}
\vspace{-2mm}
\end{figure}
\begin{figure*}[tb]
\centering
\resizebox{0.95\textwidth}{!}{
\begin{tabular}{c}
\includegraphics[width=\textwidth]{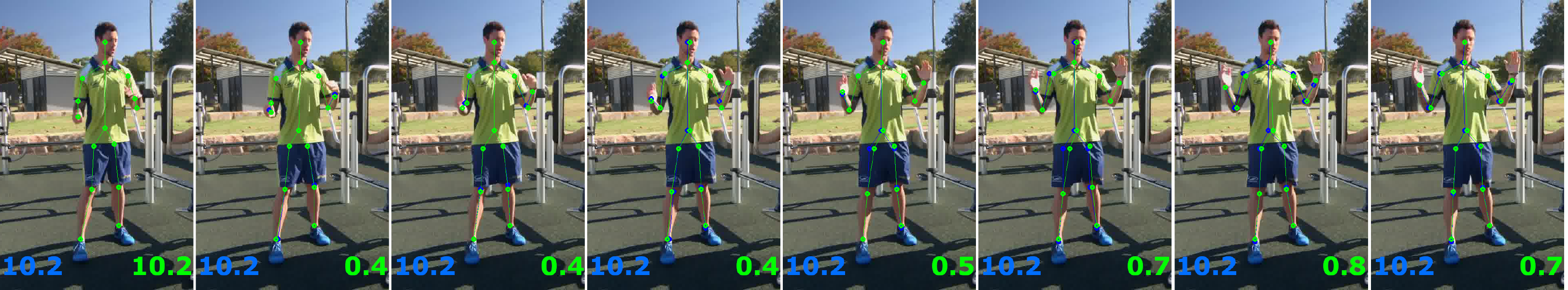}\\
\includegraphics[width=\textwidth]{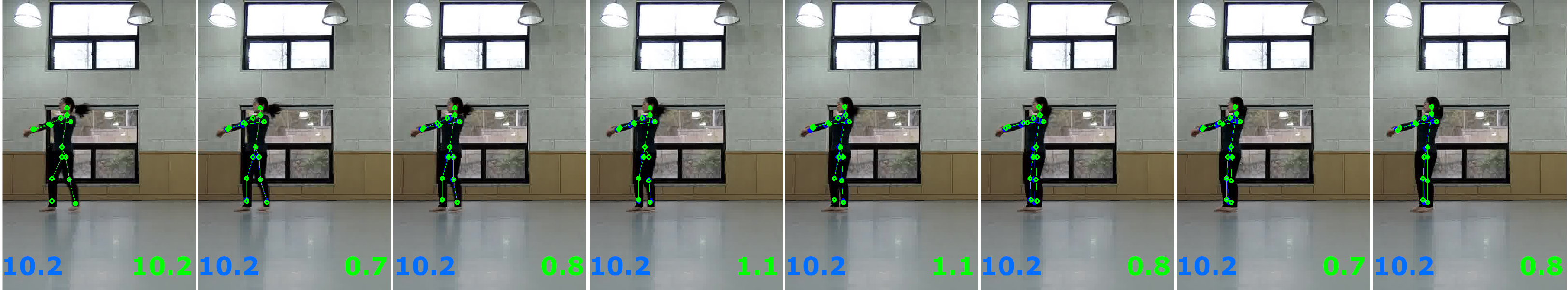}
\end{tabular}}
\vspace{-2mm}
\caption{Qualitative comparison between \textcolor{blue}{HRNet} and \textcolor{green}{HRNet+Skip-Conv}. GMAC count is reported for each frame. \skipconv significantly reduces the computations with minimal difference in predictions.
Frames from~\cite{cc5,cc6}.
}
\label{fig:pose_macs}
\vspace{-5mm}
\end{figure*}
\paragraph{Comparison to state of the art}
We compare~\skipconv to two categories of prior works: \emph{i)} task specific methods, which are dedicated to efficient human pose estimation in video~\ie by dynamic kernel distillation (DKD)~\cite{nie2019dynamic}. \emph{ii)} task agnostic methods, which optimize the model efficiency for any task and architecture,~\ie by model compression and pruning. For this purpose, we apply $\ell_0$ channel pruning~\cite{louizos2018learning}, Spatial SVD (S-SVD)~\cite{jaderberg2014speeding}, and Weight SVD (W-SVD)~\cite{wsvd} to compress HRNet-w32 at different efficiency vs.~accuracy trade-offs. For S-SVD and W-SVD, we use greedy search to select the optimal rank per layer as implemented in~\cite{aimet}. For batch-norm layers in ~$\ell_0$ channel pruning, we estimate the statistics during test on a large batch of $48$ images, as it performs better in our experiments than using batch statistics from training. Finally, we compare to a conditional computation baseline on images, $CC-image$, by applying Gumbel gates to the raw frames, instead of residual frames, similar to~\cite{verelst2020dynamic}.

\begin{table}[b]
\centering
\resizebox{\columnwidth}{!}{
\begin{tabular}{lcccc} 
\toprule
                           & \textbf{GMAC} & \textbf{Time (ms)} & \textbf{MAC Red.} & \textbf{Time Red.}  \\ 
\midrule
\textbf{Conv}                       & 10.19   & 548       & 1.00 $\times$    & 1.00 $\times$      \\
\midrule
\multirow{3}{*}{\textbf{Skip-Conv}} & 4.07    & 369       & 2.51 $\times$    & 1.48 $\times$      \\
                           & 2.35    & 287       & 4.33 $\times$     & 1.91 $\times$       \\
                           & 1.29    & 134       & 7.92 $\times$    & 4.09 $\times$      \\
\bottomrule
\end{tabular}}
\caption{MAC count vs runtime reductions on a HRNet-w32 architecture. The MAC count reductions obtained by~\skipconv translate to wall-clock runtimes.}
\label{tab:runtime}
\vspace{-5mm}
\end{table}
Table~\ref{tab:pose_estimation_sota} reports the comparison with the state of the art models on the JHMDB dataset.
By comparing \skipconv results with the backbone network HRNet-w32, results highlight that skipping redundant computations allows a reduction in MAC count by roughly a factor of $2$, even with a remarkable improvement in PCK from $94.5$ to $95.1$.
We attribute such a performance increase to a regularizing effect from the firing of stochastic gates during training.
Moreover, when compared with DKD~\cite{nie2019dynamic}, \skipconv yields again a $1$ point margin in PCK, with a relative cost reduction of $38.7\%$. 
Finally, out of the model compression baselines, S-SVD excels by halving the MAC count of HRNet-w32 with a minimal reduction in accuracy, even outperforming DKD in terms of the PCK vs cost trade-off.
Notably, W-SVD and L0 regularization achieve similar compression rates, but with more severe performance degradations.

The comparison between \skipconv and model compression baselines can be best understood by looking at Figure~\ref{fig:plot_jhmdb}, that reports PCK and MAC count at different operating points. 
The figure clearly shows the better trade-off achieved by \skipconvdot, which is able to retain the original HRNet-w32 performance whilst reducing the cost by more than a factor $4$.
On the contrary, other baselines experience higher drop in performance when increasing their compression ratios, with the best trade-off achieved by S-SVD.
However, we remark that model compression and \skipconv tackle two very different sources of inefficiency in the base model:
if the former typically focuses on cross-channel or filter redundancies, the latter tackles temporal redundancies.
For these reasons, a combination of the two approaches could further improve efficiency, as also testified by the cyan line in Figure~\ref{fig:plot_jhmdb}, that we obtain by applying \skipconv to different S-SVD compressed models.
Indeed, the combination of such strategies outperforms both of them, especially in the low-cost regime.
Finally, the comparison between ~\skipconv and $CC-image$ highlights the importance of conditioning on residuals, as they provide a strong prior to distinguish relevant and irrelevant locations.
Figure~\ref{fig:pose_macs} depicts examples of \skipconv predictions.
\paragraph{Runtime speed up}
We investigate how the theoretical speed ups, measured by MAC count reductions, translate to actual wall clock runtimes. Following~\cite{dong2017more} we use \emph{im2col} based implementation of sparse convolutions. This algorithm reformulates the convolution as a matrix multiplication between input tensor and convolution kernels flattened as two matrices. The multiplication is computed only on non-sparse columns while filling the other columns by zero. We report the overall wall clock time spent on conv layers vs ~\skipconv layers for a HRNet-w32 architecture. The runtimes are reported on CPU\footnote{Intel Xeon e5-1620 @ 3.50GHz.}.
As reported in Table~\ref{tab:runtime}, the MAC count reductions obtained by ~\skipconv translate to wall clock runtimes. The improvements on runtimes are roughly half of the theoretical speed ups as MAC count does not count for memory overheads involved in sparse convolutions. The gap between theoretical and real runtime improvements can be further reduced through highly optimized CUDA kernels as demonstrated in ~\cite{ren2018sbnet, verelst2020dynamic}.
\subsection{Ablation studies}
\label{sec:ablation}
\paragraph{Impact of gating function}
\begin{figure}[tb]
\centering
\includegraphics[width=0.9\columnwidth]{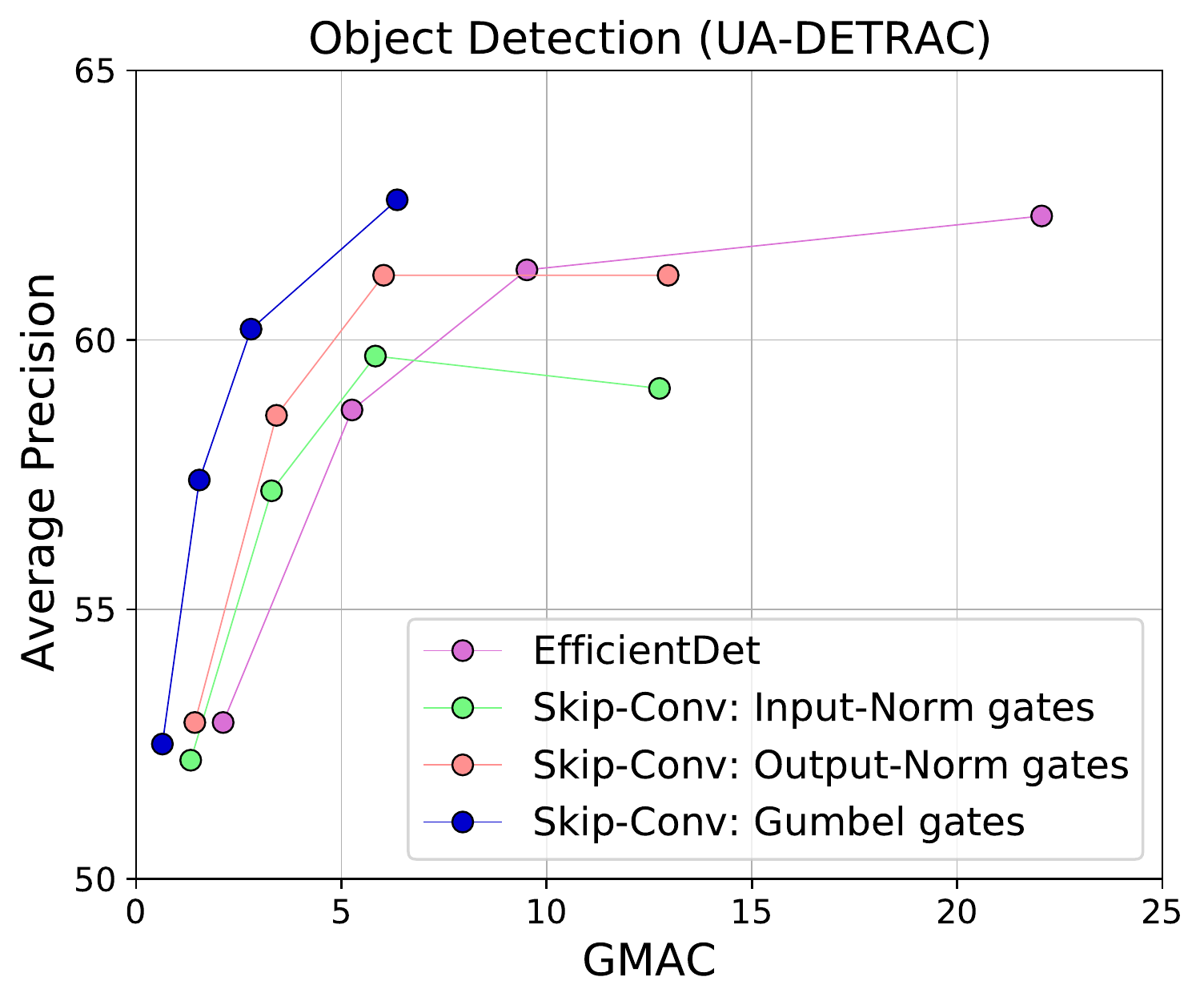}
\vspace{-2mm}
\caption{Comparison of different gates for~\skipconvdot. Output-Norm gates improve EfficientDet, though not as effectively as Gumbel gates.}
\label{fig:gates_object_detection}
\end{figure}
\begin{table}[b]
\bgroup
\newcolumntype{Y}{>{\centering\arraybackslash}X}
\centering
\resizebox{0.9\columnwidth}{!}{
\begin{tabularx}{\columnwidth}{YYY}
\toprule
\textbf{Block size}& \textbf{GMAC} & \textbf{PCK}\\
\midrule
\textbf{$1\times 1$} & 5.06 & 95.0\\
\textbf{$2\times 2$} & 2.43 & 94.5\\
\textbf{$4\times 4$} & 2.98 & 94.9\\
\textbf{$8\times 8$} & 4.18 & 95.0\\
\bottomrule
\end{tabularx}}
\caption{Impact of structured gates on pose estimation. Structured gates perform comparable to unstructured gates ($1 \times 1$ blocks), while allowing for efficient implementations.}
\label{tab:structure}
\egroup
\vspace{-5mm}
\end{table}
We study the impact of gating on~\skipconv by evaluating EfficientDet architectures using three different gate functions (Section~\ref{sec:method_gating}): \emph{i)} Input-Norm gates with threshold value $\epsilon=1e-2$; \emph{ii)} Output-Norm gates, with threshold value $\epsilon=15e-5$; \emph{iii)} Gumbel gates are trained with sparsity coefficient $\beta=1e-2$.

Figure~\ref{fig:gates_object_detection} illustrates that Gumbel gates outperform both the Input-Norm and Output-Norm gates. 
This behavior is expected, as Gumbel gates are trained end-to-end with the model and they learn to skip the residuals, regardless of their magnitude, if it does not affect the task loss. 
Therefore, they effectively skip the big changes in background, which leads to higher computational efficiency.
Moreover, we observe that output-norm gates outperform input-norm gates as they rely on a more precise approximation involving weight norms.
Despite their simplicity, output-norm gates improve the efficiency of EfficientDet with reasonable accuracy drop. 
As an example, for D2 configuration output-norm gates reduce the cost from $9.52$ to $6.03$ GMAC with a similar AP of $61.2$.
Although output-norm gates are less effective than Gumbel gates, they are practically valuable as they can be plugged into any trained network without any labeled video or training required.

\paragraph{Structured gating}
We experiment with structured gating on JHMDB (split 1) and report results in Table~\ref{tab:structure}.
The table reports the accuracy and efficiency of two unstructured models trained with different sparsity coefficients $\beta$, along with structured models with $4\times4$ and $8\times8$ blocks.
It can be noted how adding structure to the gates does not negatively impact the model, yielding results that are inline with unstructured counterparts. 
This finding suggests that structured gates introduce hardware friendliness without hurting the accuracy/cost tradeoff.
\vspace{-2mm}
\paragraph{Impact of clip length}
\begin{table}[tb]
\centering
\resizebox{0.95\columnwidth}{!}{
\begin{tabular}{lcccccc}
\toprule
&& \multicolumn{2}{c}{$T_{train}=4$} && \multicolumn{2}{c}{$T_{train}=8$}\\
\cmidrule{2-7}
&& \textbf{PCK} & \textbf{GMAC} && \textbf{PCK} & \textbf{GMAC}\\
\midrule
$T_{test}=4$ && 95.3 & 3.10  && 94.5 & 3.56\\
$T_{test}=8$ && 94.3 & 1.91  && 94.5 & 2.43\\
$T_{test}=\infty$ && 89.3 & 1.18  && 94.2 & 1.80\\
\bottomrule
\end{tabular}}
\caption{Results for pose estimation when training and testing with different clip lengths $T$.}
\label{tab:clip_length}
\vspace{-3mm}
\end{table}
We study the sensitivity of \skipconv to clip length used during training and to reference frame reset frequency during test.
Table~\ref{tab:clip_length} shows results on JHMDB (split 1), where we train Gumbel gates with clips of $4$ or $8$ frames with a sparsity factor $\beta=5e-5$.
Similarly, we instantiate a new reference frame during test every  $4$ or $8$ frames, or even only once at the beginning of each sequence ($t=\infty$).
As one can expect, the table shows how decreasing the number of expensive reference frames improves efficiency.
This comes, however, at a minor cost in PCK, with a drop of $0.3$ PCK for processing up to $40$ frames sequences when training with clips having length $T=8$.
\section{Conclusion}
We propose Skip Convolutions to speed up convolutional nets on videos.
Our core contribution is the shift of the convolution from the content frames to the residual frames, both at input and intermediate layers.
Operating on residual frames allows to skip most of the regions in the feature maps, for which representations can simply be copied from the past.
We further encourage this regime by per layer gating functions, for which we propose several trainable and off-the-shelf designs.

As a potential limitation, we highlight it is unclear how our model would perform in the presence of severe camera motion. 
In such situations, residual frames wouldn't bear that much information about relevant regions, thus a higher burden would be put on the gating function.
Coupling~\skipconv with learnable warping functions helps compensating for severe camera motions, and is deferred to future work.
\paragraph{Acknowledgements} We thank Max Welling, Fatih Porikli and Arash Behboodi for feedback and discussions.
\clearpage
{\small
\balance
\bibliographystyle{ieee}
\bibliography{egbib}
}
\clearpage
\twocolumn[
\Large
\begin{center}
\textbf{Skip-Convolutions for Efficient Video Processing}\\
\textbf{Supplementary material}\vspace{1cm}
\end{center}
]
\section{Experiments on video classification}
\begin{table}[t]
\centering
\resizebox{0.9\columnwidth}{!}{
\begin{tabular}{lcc} 
\toprule
   & \textbf{Accuracy} & \textbf{GMAC} \\ 
\midrule
\textbf{TSN-25} & 55.49   & 194.98 \\
\textbf{TSN-25 + \skipconv} & 54.77   & 41.02 \\
\midrule
\textbf{TSN-6} & 53.84   & 46.80 \\
\textbf{TSN-6 + \skipconv} & 53.77   & 35.83 \\
\bottomrule
\end{tabular}}
\caption{Video classification results. ~\skipconv reduces the computation cost with a minor accuracy drop.}
\label{tab:hmdb}
\end{table}
Although our work is mainly focused on stream processing, \ie video tasks where a
spatially dense prediction is required for every frame, our \skipconv model can in principle enable improvement in efficiency of classification models.
We hereby conduct a preliminary experiment with video classification on human action dataset~\cite{jhuang2013towards}, and consider the split-1 using RGB modality. We report Top-1 accuracy for center-crop inference. 
As a backbone model, we rely on Temporal Segment Networks (TSN)~\cite{tsn}, based on a ResNet-101.
We study the performance of~\skipconv in two inference setup: \emph{i.} TSN-25, where inference is carried out over 25 frames per clip (sampled uniformly from the whole video as in~\cite{tsn}). \emph{ii.} TSN-6, where inference is carried out over 6 frames per clip, so there are much less redundancies between frames.

As reported in Tab.~\ref{tab:hmdb}, ~\skipconv reduces the computation cost with a minor accuracy drop. The computation gain is higher for the high frame-rate model (TSN-25) as there are more redundancies between video frames. For the low frame-rate model (TSN-6), the compute is reduced from 46.80 to 35.83 GMACs even though the frame redundancies and residual sparsities are fairly low because of the coarse frame sampling.

Finally, we remark that many state-of-the-art video classification architectures rely on 3D backbones~\cite{feichtenhofer2020x3d,piergiovanni2019tiny,tran2018closer}.
Although the focus of this paper has been on 2D \skipconvdot s, Eq. 1 and 2 in the main paper can be extended to the case of 3D convolutions as a linear operator.
\section{Sparsity ratios}
In this section we analyze the amount of sparsity induced by \skipconv in different levels of a backbone network.
To this end, we refer to the pose estimation experiments described in Sec.~4.2 of the main paper, and we rely on the same setting by considering the JHMDB dataset~\cite{jhuang2013towards} with a HRNet-w32 backbone network~\cite{hrnet}.
We train the \skipconv model with Gumbel gates under different sparsity objectives, by varying $\beta$ in $[1e-5,5e-5,10e-5,15e-5]$.
For completeness, we also report the performance of these models, that score $[0.95, 0.94, 0.93, 0.91]$ in PCK respectively.
We then measure how the firing probability of gates in \skipconv changes at different depths in the network.

\begin{figure}[b]
\centering
\includegraphics[width=0.95\columnwidth]{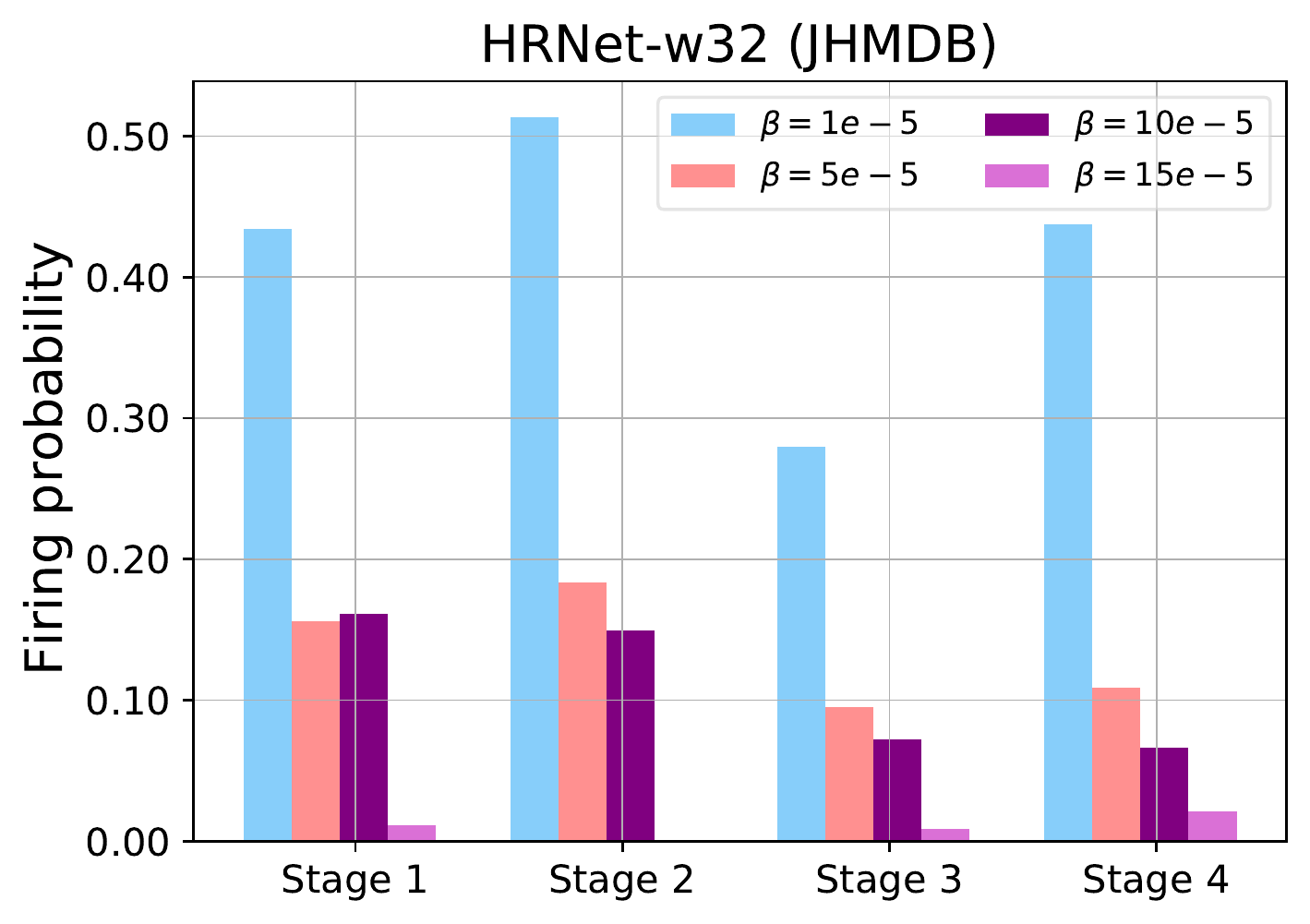}
\caption{Per-stage sparsity level of HRNet-32.}
\label{fig:sparsity_ratio}
\end{figure}

The results are summarized in Fig.~\ref{fig:sparsity_ratio}, where we report the probability of firing at different stages of the base HRNet-w32 model, averaged over all test examples, different layers within the same stage, and the three splits commonly used in pose estimation protocols.
The figure highlights how, in general, \skipconv allows to bypass a significant amount of computation.
Even under very mild sparsity constraints (\ie $\beta=1e-5$), the Gumbel gates learn to skip more than half of the pixels in feature maps overall.
For intermediate values of $\beta$, firing probabilities drop to below 0.2, and in some cases fall under 0.1 for later stages in the network.
For high sparsity coefficients (\ie $\beta=15e-5$) \skipconv mostly relies on features from the first frame in the input clip, and triggers computation very occasionally (0.8\% to 2.1\% of pixels).
Interestingly, for stage 2 has a firing probability of zero: this means that the model only relies on features from the reference frame for those layers, and that they suffice to carry out correct predictions.
\end{document}